\documentclass[journal,10pt,twocolumn,twoside]{IEEEtran}
\usepackage{amsmath,amsfonts,amssymb,amsbsy,bm,paralist,theorem,color}
\usepackage{graphicx,algorithmic,algorithm,relsize}
\usepackage{multicol}
\usepackage{multirow}
\usepackage{subfigure}
\usepackage{caption}
\usepackage{cite}
\usepackage{pifont}
\usepackage{amsfonts,amssymb}
\usepackage{bbm}

\usepackage{epstopdf} 

\DeclareMathOperator*{\st}{s.t.}

\graphicspath{{fig/}}

\definecolor{orange}{RGB}{255,107,0}
\definecolor{green}{RGB}{0,160,20}

\begin{document}
	\title{Joint Age-based Client Selection and Resource Allocation for Communication-Efficient Federated Learning over NOMA Networks}
	\author{Bibo~Wu,~\IEEEmembership{}
		Fang~Fang,~\IEEEmembership{Senior Member, IEEE,}
		Xianbin Wang,~\IEEEmembership{Fellow, IEEE}
		
		\thanks{Bibo Wu, Fang Fang and Xianbin Wang are with the Department of Electrical and Computer Engineering, and Fang Fang is also with the Department of Computer Science, Western University, London, ON N6A 3K7, Canada. (e-mail: \{bwu293, fang.fang, xianbin.wang\}@uwo.ca).}}
	
	\maketitle
	\begin{abstract}
	In federated learning (FL), distributed clients can collaboratively train a shared global model while retaining their own training data locally.
	Nevertheless, the performance of FL is often limited by the slow convergence due to poor communications links when FL is deployed over wireless networks. 
	Due to the scarceness of radio resources, it is crucial to select clients precisely and allocate communicaiton resource accurately for enhancing FL performance. 
	To address these challenges, in this paper, a joint optimization problem of client selection and resource allocation is formulated, aiming to minimize the total time consumption of each round in FL over a non-orthogonal multiple access (NOMA) enabled wireless network. 
	Specifically, considering the staleness of the local FL models, we propose an age of update (AoU) based novel client selection scheme.
	Subsequently, the closed-form expressions for resource allocation are derived by monotonicity analysis and dual decomposition method. 
	In addition, a server-side artificial neural network (ANN) is proposed to predict the FL models of clients who are not selected at each round to further improve FL performance. 
	Finally, extensive simulation results demonstrate the superior performance of the proposed schemes over FL performance, average AoU and total time consumption.
	\end{abstract}

    \begin{IEEEkeywords}
    Age of update; artificial neural network; client selection; federated learning; resource allocation.
    \end{IEEEkeywords}

	\vspace{-1em}

	\section{Introduction}
	In recent years, there has been a significant increase in both volume of the data traffic and a number of smart devices. The trend can be attributed to the rapid proliferation of Internet of Things (IoT) applications.
	This necessitates data-intensive machine learning (ML) to effectively utilize the massive amount of data for supporting intelligent services and applications \cite{FLIoT,FLmag}.
	However, traditional ML relies on centralized data processing, which requires all distributed raw data (normally with large size) to be gathered for the ML training \cite{DLmag,MLmag}.
	Thus, such centralized paradigm inevitably brings heavy communication overhead and growing concerns on data privacy.
    To effectively address these issues, various decentralized learning approaches including federated learning (FL) have been investigated extensively in recent years.
    In comparison with centralized machine learning, FL allows edge entities, namely clients, to collaborate on training a common global model while retaining their own raw data locally \cite{FL6G, WFL}.
    As a direct benefit, distributed computing resources can be effectively utilized without incurring data privacy concerns.
    Specifically, clients in FL utilize the local private data to train their local FL models which they then send to the server.
    Subsequently, after receiving these local models successfully, the server performs model aggregation to train a global FL model that is returned to all clients.
    Through multiple rounds of learning and communication, the global FL model converges while eliminating the need to collect the private data from the involved clients. 
    As a result, this new approach overcomes both the data privacy concerns and high communication overhead challenge in conventional  centralized ML \cite{FLsurvey}.
    
    Client selection plays an important role in FL, given that all clients involved will determine the overall learning performance. 
    Generally, the number of selected clients at different rounds of FL is limited due to communication and computing resource constraint \cite{FLcom}.
	Most of the existing FL client selection policies are designed mainly based on threshold or probability \cite{CSsurvey}.
	However, timely model updates are crucial for the performance of FL as staleness of model update can significantly increase the convergence time \cite{AoI}.
	To address this issue, the concept of age of update (AoU), is adopted as the metric to measure the elapsed time since the server received the latest update in FL and further to assist client selection \cite{AoImag}. 
	Note that if the AoU value of a client is large, the degree of staleness associated with that client's local FL model will also be large, which negatively affects the FL convergence.
	In this case, the server needs to obtain timely FL training updates from clients.
	
	The FL outcome can be improved by enhancing the performance of underlying wireless communication networks for reduced access latency and increased throughput. 
	However, this could be a challenge when the overall communication resource is insufficient. 
	As a result, resource-efficient channel access techniques could be very helpful in improving the performance of FL.
	With the ever-increasing level of resource scarceness, it has been recognized that non-orthogonal multiple access (NOMA) is an important enabler to enhance communication efficiency in FL systems \cite{Ding_Survey_2017, Liu_Mag_2017}.
	It has also been proved that NOMA can achieve positive effect on the decrease of AoU by improving channel access and reducing user collisions \cite{AoINOMA1}.
	When integrating NOMA in FL systems, multiple clients can transmit their signals simultaneously on the same resource block, hence achieving the low level AoU of the entire FL system.
	Thus, the integration of AoU and NOMA to achieve efficient resource allocation in FL is deserved to be investigated.

	\subsection{Related Literature}
	To improve FL performance, extensive research has been conducted in related literature, including client selection and resource allocation in FL systems \cite{Konecn2016FederatedOD}.
	
	\begin{itemize}
		\item [1)] Client selection in FL: 
		It has been shown that client selection is crucial in determining the FL performance, particularly in scenarios where there are numerous clients and limited wireless resources \cite{CSsurvey}. 
		Generally, FL is characterized by dynamic environments and limited network bandwidth, which makes the number of clients engaging in global FL model training at each round is limited.
		Therefore, the proper selection policy is vital to achieving the desired accuracy and convergence in FL.
		Current existing FL client selection policies are designed mainly based on probability \cite{FLCS3}, threshold \cite{FLCS1, FLCS2}, reputation \cite{FLCS4,FLCS5} and so forth.
		More specifically, the authors in \cite{FLCS3} selected the clients in FL via a probabilistic policy, aiming to solve the minimization problem of communication time.	
		In \cite{FLCS1}, the client that consumes the least amount of updating time was selected at each round by the optimal bandwidth allocation algorithm.
		\cite{FLCS2} presented a threshold-based client selection approach, which is designed to choose as many clients as possible who are capable of completing the local FL training within a given threshold to achieve the desired performance.
		The authors in \cite{FLCS4,FLCS5} proposed a reputation-based client scheduling strategy, which selects the trustworthy clients based on both direct interactions and recommendations from other servers.
		Nevertheless, the concept of AoU was not considered in these schemes, which plays a crucial role in determining the fast FL convergence.
		
		Realizing the significance of AoU to the global model update in FL systems, an AoU based client selection scheme was proposed in \cite{FLage} to achieve the global loss minimization.
		According to \cite{AoUcon}, an AoU based client scheduling policy was proposed that accounts for both the staleness of local models and the channel qualities to increase FL efficiency.
		Nevertheless, NOMA was not considered in these works, which can further improve the performance of FL with enhanced communication efficiency.

		\item [2)] Resource allocation in FL:
		A number of works have proposed resource allocation in FL with the goal of minimizing FL convergence time \cite{FLT1, FLT3,FLT4}.
		Specifically, aiming to achieve the minimization of FL convergence time, a joint problem of client scheduling and wireless resource allocation was formulated in \cite{FLT1, FLT3}.
		In \cite{FLT4}, to address the total FL delay minimization problem, the authors empirically studied the impacts of local computation and model compression on FL delay, communication and computing.
		Besides, the resource allocation of energy efficient FL was studied in \cite{FL1, FL3}.
		In particular, with the objective of total energy consumption minimization, the authors in \cite{FL1} jointly considered the allocation of dataset size, transmitting power and subcarrier in FL systems.
		In \cite{FL3}, the problem of computational resource allocation and energy efficient transmission in FL was investigated, and the closed-form solutions were derived via an iterative algorithm.
		There also are some other works investigating the global FL loss minimization problem while guaranteeing efficient resource allocation \cite{FLT2, FLloss1, FLloss2}.
		The authors in \cite{FLT2} jointly addressed the client selection and bandwidth optimization problems for the purpose of minimizing the FL training loss.
		It was proposed to use the packet error rate in \cite{FLloss1} to determine the parameter transmission, followed by a joint analysis of resource allocation and client selection.
		In \cite{FLloss2}, aiming to achieve the global loss minimization, FL was deployed in a massive multiple-input-multiple-output (MIMO) system to jointly optimize the client selection strategy and power allocation.
		However, all these works consider the orthogonal multiple access (OMA) technology in their FL systems, which has limited improvement to the FL performance in resource constrained scenarios.
		
		When integrating NOMA technology in FL, the same resource block can be shared by multiple clients to simultaneously upload their respective local models to the server, which improves the data rate and enhances FL performance \cite{FLNOMA1, FLNOMA3, FLNOMA4}.
		In \cite{FLNOMA1}, the authors proposed a FL system over NOMA network to realize the maximization of sum transmission rate with optimal power allocation.
		In \cite{FLNOMA3}, an intelligent reflecting surface aided over-the-air FL system over NOMA network was proposed. 
		By optimizing the reflection coefficients and transmitting power, the achievable data rate was maximized.
		According to \cite{FLNOMA4}, NOMA was combined with adaptive gradient sparsification and quantization in order to accelerate FL updates in uplink channels while taking spectrum limitations into account.

	\end{itemize}

	\subsection{Motivations and Contributions}
	From above observations, the client selection and resource allocation are two crucial factors needed to be considered jointly in determining the performance of FL, especially in resource constrained systems.
	It is nessary to consider staleness of model update in FL, which significantly affect the convergence performance of FL \cite{AoI}.
	Therefore, AoU should be regarded as a metric for the protocol of client selection, that is, there is a tendency for the server to select clients with larger AoU for model aggregation at any FL round.
	In this case, the fairness of client selection can be guaranteed, hence mitigating the overfitting risk of global FL model.
	There are some current works considering AoU in FL systems with OMA transmission, while ignoring the effect of NOMA on AoU \cite{AoImag, AoUcon}.
	With NOMA in FL, more clients can be selected by the server for model aggregation at each round, hence achieving the low level AoU of the entire system \cite{AoINOMA1}.
	Besides, NOMA is promising to improve communciation efficiency in resource constrained scenarios, which is beneficial to facilitate the implementation of FL.
	Therefore, the integration of AoU and NOMA in FL is necessary, but little research has been done in this regard.
	
	Motivated from above analysis, we propose an AoU based client selection and optimal resource allocation scheme to minimize the convergence time in NOMA enabled FL system.
	Specifically, we formulate a joint client selection and resource allocation problem, which is a non-convex mixed-integer non-linear programming (MINLP) problem.
	Subsequently, it is decomposed into two subproblems, including the client selection subproblem and the resource allocation subproblem.
	Then, a novel AoU based client selection scheme is proposed and we obtain the closed-form expressions for resource allocation.
	To make further improvement of FL performance, we propose to deploy the server-side artificial neural network (ANN) to realize the prediction of the unselected clients' local models at every round.
	The following are the main contributions of this paper:
	\begin{enumerate}[1)]
		\item We propose an AoU based FL over NOMA system, where the AoU's impact on FL performance is analyzed. 
		The uplink NOMA transmission can not only decrease the AoU of the entire system but also increase the number of clients who are selected by the server at each FL round.
		The joint optimization problem of client selection and resource allocation is formulated to achieve the minimization of total time consumption of each FL round.
		We decompose it into two subproblems, including the client selection subproblem as well as the resource allocation subproblem.
		
		\item For the AoU based client selection design, an upper bound is derived for the convergence of the AoU-based global FL model. 
		Based on this, we formulate a weighted client selection problem, and obtain a selecting list of clients based on their AoU and data size. 
		For the resource allocation subproblem, given the AoU based client selection scheme, the monotonicity analysis and dual decomposition method are utilized to derive the closed-form expressions for computational resource allocation and power allocation, respectively.
		
		\item To make further improvement of FL performance, we propose to deploy the server-side ANN to obtain the predicting results of the local FL models of clients who are not selected at every FL round because of the restricted communication resources. 
		With the deployment of ANN, there is an increase for the number of clients engaging in model aggregation at the server, resulting in an improvement of FL performance without extra communication overhead.
		
	\end{enumerate} 
	
	\subsection{Organization}
	The rest of this paper is organized as follows.
	We propose an AoU based FL over NOMA system and formulate the total time consumption minimization problem in Section II.
	In Section III, we propose an AoU based client selection scheme and derive the optimal solutions for resource allocation.
	In Section IV, the deployment of ANN to derive the predicting results of the unselected clients' local FL models is proposed.
	Section V shows the extensive simulation results and corresponding analysis.
	We conclude this paper in Section VI.

	\section{System Model and Problem Formulation}
	\begin{figure}[t] 
		\includegraphics[width=0.5\textwidth]{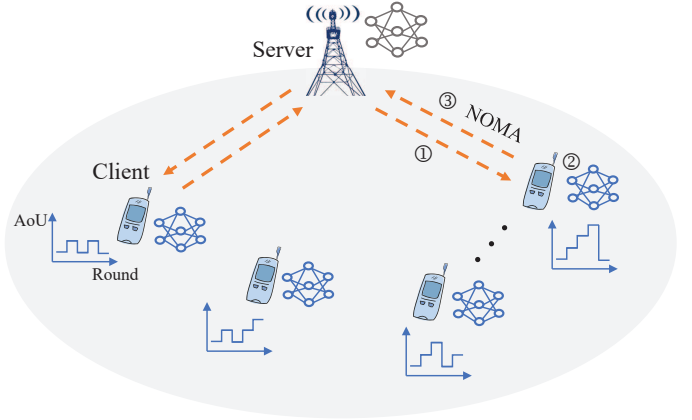}
		\centering
		\caption{FL over NOMA network.}
		\label{System_model}
	\end{figure}

	\subsection{AoU based FL Model}
	Consider a NOMA based FL system, where a wireless server cordinates $N$ clients denoted as ${{\cal{N}}} = \left\{ {1,2,...,{N}} \right\}$ to train a FL model collaboratively through uplink NOMA transmission, as shown in Fig. \ref{System_model}. 
	Both the server and the client are assumed to be equipped with single antenna.
	We summarize the entail training process of FL as:
	At any FL training round defined as $t$, the server sends the global FL model ${{\mathbf{w}}^t}$ to the clients. 
	After receiving the global model ${{\mathbf{w}}^t}$, any selected client $n \in \cal {N}$ performs the local model training using its private data.
	Note that the number of selected clients at each round is limited to reduce the communication overhead within limited wireless communication resources. 
	Then, these trained local FL models will be transmitted to the server by the selected clients for model aggregation via uplink NOMA network.
	After many rounds of communicating and training, the global FL model comes to convergence finally.
	In this paper, it is assumed that the server is capable of providing reliable and lossless communications for downlink broadcast channels.
	Thus, the communication cost for downlink transmission is negligible \cite{FLassum}.
	
	The AoU can describe the staleness of model update in FL, and it has been proved that large staleness of model update will negatively impact the learning convergence in FL systems \cite{AoI}.
	Thus, to accelerate FL convergence, we propose an AoU based client selection policy in this paper, so that the AoU of the whole system can be kept at a low level, i.e., the model update in FL is guaranteed as fresh as possible.
	We define the client $n$'s AoU at round $t$ as $A_n^t$, which can be calculated by \cite{AoIdef}
	\begin{equation}\label{AoU}
		\begin{aligned}
			A_n^t = \left\{ {\begin{array}{*{20}{l}}
					{A_n^{t - 1} + 1,{\text{ }}S_n^{t - 1} = 0}, \\ 
					{1,{\text{ }}S_n^{t - 1} = 1}, 
			\end{array}} \right.
		\end{aligned}
	\end{equation}	
	where the binary variable $S_n^t \in \left\{ {0,1} \right\}$ denotes the selecting status of client $n$, i.e., $S_n^t=1$ indicates client $n$ is selected for model aggregation at round $t$; $S_n^t=0$ otherwise.
	Assume there are at most $N^t$ clients selected at round $t$ and define the set of selected clients as ${{\cal{N}}^t} = \left\{ {1,2,...,{N^t}} \right\}$.
	We can describe the essence of \eqref{AoU} as: If client $n$ was selected at last round, its AoU $A_n^t$ will be reset to $1$; otherwise its $A_n^t$ will add $1$.
	To keep the AoU of the entire system at a low level, the server inclines to select the clients with larger AoU at each round.
	In orther words, the larger AoU can reflect more informative update, which benefits the convergence of FL.
	To prioritise selecting clients with higher AoU for aggregation, define $a_n^t$ as a weighting factor:
	\begin{equation}\label{a_n}
		\begin{aligned}
			a_n^t = \frac{{A_n^t}}{{\sum\nolimits_i^N {A_i^t} }}.
		\end{aligned}
	\end{equation}

	In FL systems, the local FL model of any client $n$ is given by
	\begin{equation}
		\begin{aligned}
			{F_n}\left( {{\mathbf{w}}^t} \right) = \frac{1}{{{\beta _n}}}\sum\limits_{i = 1}^{{\beta _n}} {l\left( {{\mathbf{w}}^t;{\mathbf{x}}_{n,i}^t,y_{n,i}^t} \right)},
		\end{aligned}
	\end{equation}
	where ${\beta _n}$ represents the number of data samples of client $n$, ${l\left( {{\mathbf{w}}^t;{\mathbf{x}}_{n,i}^t,y_{n,i}^t} \right)}$ denotes client $n$'s local loss function, ${\mathbf{x}}_{n,i}^t$ and $y_{n,i}^t$ are the $i$-th input data with ${i \in \left\{1,...,\beta_n \right\}}$ and the corresponding label, respectively.
	Integrating the proposed AoU based client selecting weight in model aggregation, we can present the global FL model at the server as 
	\begin{equation} \label{global}
		\begin{aligned}
			F\left( {{{\mathbf{w}}^t}} \right) = \frac{{\sum\nolimits_{n = 1}^N {a_n^{t-1}S_n^{t-1}{\beta _n}} {F_n}\left( {{\mathbf{w}}^{t-1}} \right)}}{{\sum\nolimits_{n = 1}^N {a_n^{t-1}S_n^{t-1}{\beta _n}} }},
		\end{aligned}
	\end{equation}	
	which is defined as an averaging aggregation manner.
	We can observe from \eqref{global} that the local FL models from the selected clients with larger AoU holds larger weight for the model aggregation, which reflects the impact of AoU on the global FL model.

	\subsection{Time Consumption Models}
	In this paper, the total time consumption at any FL round is divided into the training time of local FL model and the uplink NOMA transmission time.
	For any selected client $n$, after receiving the global FL model, its local model will be trained by utilizing the equipped CPU resources.
	Define $f_n$ as the client $n$'s CPU frequency, and the local training time consumption can be calculated by
	\begin{equation}\label{}
		\begin{aligned}
			T_n^{cp} = \frac{{\mu {\beta _n}}}{{{\tau _n}{f_n}}},
		\end{aligned}
	\end{equation}
	where $\mu$ is the coefficient of computational resource allocation indicating the required number of CPU cycles to train one sample, and ${\tau _n}$ denotes a designed proportion of computational frequency, which meets ${\tau _n} \in \left[ {0, 1} \right]$.

	The application of NOMA in FL not only improves the spectrum efficiency via simultaneous transmission of clients over the same resource block, but also has a positive effect on the decrease of AoU with more selected clients at each FL round compared with OMA transmission, thus contributing to the improvement of FL performance.
	In our proposed FL over NOMA system, the selected clients transmit their local FL models through uplink NOMA transmission over a same channel.
	Denote $p_n^t$ and $h_n^t$ as the transmitting power of client $n$ and the channel gain between the server and client $n$ at round $t$, respectively.
	Let $x_n^t$ be the client $n$'s transmitted symbols at round $t$, which can be normalized as $\left\| {x_n^t} \right\|_2^2 = 1$.
	With NOMA transmission, the received signal of the server is the superposition of the signals transmitted from the selected clients, which can be presented as 
	\begin{equation}\label{}
		\begin{aligned}
			{y^t} = \sum\limits_{n = 1}^{{N^t}} {\sqrt {p_n^t} h_n^tx_n^t}  + {n^t},
		\end{aligned}
	\end{equation}	
	where ${n^t} \sim {\cal{CN}} \left( {0,{\sigma ^2}} \right)$ represents the additive  white Gaussian noise (AWGN).
	
	To decode the signal of any specific client $n$ from the superposition of the transmitted signals, the server performs the successive interference cancellation (SIC) in NOMA systems.
	Specifically, the strongest signal is first decoded by the server, while other signals are treated as interference.
	When decoding is completed successfully, the server conducts the subtraction of the decoded signal from the superposition before decoding the next strongest signal.
	The server continues to decode the signals until all signals have been decoded.
	In this paper, without loss of generality, we make the assumption that ${p_1^t}{\left| {h_1^t} \right|}^2 > {p_2^t}{\left| {h_2^t} \right|}^2 >  \cdots  > {p_{{N^t}}^t}{\left| {h_{N^t}^t} \right|}^2$.
	Thus, client $1$'s signal is decoded firstly at the server, while the signal of client $ N^t$ is the last one.
	
	From the Shannon capacity formula, for any client $n \in \left\{ {1,2,...,{N^t} - 1} \right\}$, we can express the achievable data rate at round $t$ as
	\begin{equation}\label{}
		\begin{aligned}
			R_n^t = B{\log _2}\left( {1 + \frac{{p_n^t{{\left| {h_n^t} \right|}^2}}}{{\sum\nolimits_{j = n + 1}^{{N^t}} {p_j^t{{\left| {h_j^t} \right|}^2}}  + {\sigma ^2}}}} \right),
		\end{aligned}
	\end{equation}		
	where $B $ is the bandwidth of the channel and $\sum\nolimits_{j = n + 1}^{{N^t}} {p_j^t{{\left| {h_j^t} \right|}^2}}$ denotes the interference of the cilents decoded behind after client $n$.
	Particularly, for the client $N^t$, because there dose not exits interference from other clients, its achievable data rate at round $t$ can be written as $R_{{N^t}}^t = B{\log _2}\left( {1 + \frac{{p_{{N^t}}^t{{\left| {h_{{N^t}}^t} \right|}^2}}}{{{\sigma ^2}}}} \right)$.
	Given the above achievable data rate, the time required for uplink NOMA transmission could be expressed as:
	\begin{equation}\label{}
		\begin{aligned}
			T_n^{cm} = \frac{{{d_n}}}{{R_n^t}}, \forall n \in {\cal{N}}^t,
		\end{aligned}
	\end{equation}
	where $d_n$ is client $n$'s local model size.
	Given that local FL models consist of similar numbers of elements, it is reasonable to assume that all local FL models are of the same size, i.e., $d = d_n, \forall n \in {\cal{N}}^t$.

	\subsection{Problem Formulation}
	In our proposed AoU based FL over NOMA system, our objective is the minimization of total time consumption at each FL round under the constraints of transmission quality and maximum transmitting power, which can be given by:
	\begin{subequations}\label{t_pro}
		\begin{align}
			\mathop {\min }\limits_{{\mathbf{S}},{\mathbf{p}},{\mathbf{\tau }}}  \quad &
			\mathop {\max }\limits_{n \in {\cal{N}}} \left\{ {S_n^t\left( {T_n^{cp} + T_n^{cm}} \right)} \right\}  \label{T_fun}\\
			\st\ \quad  & \left( {R_n^t - {R_s}} \right)S_n^t \geqslant 0,  \label{QoS}\\
			& {\tau _n} \in \left[ {0, 1} \right], \label{tau} \\
			& {p_n^t} \in \left[ {0, {p_n^\text{max}}} \right], \label{pmax}\\
			& S_n^t \in \left\{ {0,1} \right\},\forall n \in {\cal{N}}, \label{Sn}
		\end{align}
	\end{subequations}
	where ${\mathbf{S}}$, ${\mathbf{p}}$ and ${\mathbf{\tau}}$ are the collections of client selection indicator, power allocation and computational resource allocation, respectively. 
	${R_s}$ is the minimum requirement of achievable data rate which guarantees the quality of access channel, and ${p_n^\text{max}}$ denotes the maximum transmitting power of client $n$. 
	Thus, \eqref{QoS} is the transmission quality assurance; \eqref{tau} and \eqref{pmax} are the range limitations for computational resource allocation coefficient and transmitting power of each client, respectively.
	It is worthwhile to notice that the presence of max in \eqref{T_fun} indicates that the model aggregation performed at the server is synchronous.
	Thus, the server conducts the model aggregation when the last local FL model is received by the server successfully at each round.

	\section{AoU based client selection and optimal resource allocation}
	Problem \eqref{t_pro} is a MINLP problem, and it is difficult to solve it optimally.
	In this section, to efficiently address the original formulated problem, we propose to decompose it into two subproblems, namely the client selection subproblem and the resource allocation subproblem.
	More specifically, we first derive the AoU based client selection scheme according to the analysis of the convergence of global FL model.
	Then, given the AoU based client selection scheme, the problem \eqref{t_pro} can be transformed into a more tractable one, and the monotonicity analysis and the dual decomposition methods can be utilized to obtain the optimal closed-form solutions of the resource allocation subproblem.

	\subsection{Design of AoU based Client Selection}
	In this paper, we assume that clients employ the full gradient descent method to train their local FL models at any given round $t$.
	Specifically, given the global FL model, the local model of client $n$ is updated according to
	\begin{equation}\label{}
		\begin{aligned}
			{F_n}\left( {{\mathbf{w}}^t} \right) = F\left( {{{\mathbf{w}}^t}} \right) - \frac{\lambda }{{{\beta _n}}}\sum\limits_{i = 1}^{{\beta _n}} {\nabla l\left( {{\mathbf{w}}^t;{\mathbf{x}}_{n,i}^t,y_{n,i}^t} \right)},
		\end{aligned}
	\end{equation}
	where $\lambda$ denotes the learning rate.
	Then, to achieve the model aggregation, all selected clients transmit their respective local models to the server. 
	Based on \eqref{global}, at round $t $, the global FL model updates as follows:
	\begin{equation}\label{}
		\begin{aligned}
			F\left( {{{\mathbf{w}}^{t + 1}}} \right) & = \frac{{\sum\nolimits_{n = 1}^N {a_n^tS_n^t{\beta _n}} {F_n}\left( {{{\mathbf{w}}^t}} \right)}}{{\sum\nolimits_{n = 1}^N {a_n^tS_n^t{\beta _n}} }}  \\
			&	= F\left( {{{\mathbf{w}}^t}} \right)  
			- \frac{{\lambda \sum\nolimits_{n = 1}^N {a_n^tS_n^t} \sum\nolimits_{i = 1}^{{\beta _n}} {\nabla l\left( {{\mathbf{w}}^t;{\mathbf{x}}_{n,i}^t,y_{n,i}^t} \right)} }}{{\sum\nolimits_{n = 1}^N {a_n^tS_n^t{\beta _n}} }} \\ 
			&	= F\left( {{{\mathbf{w}}^t}} \right) - \lambda \left[ {\nabla G\left( {{{\mathbf{w}}^t}} \right) - \hat F\left( {{{\mathbf{w}}^t}} \right)} \right], \\ 
		\end{aligned}
	\end{equation}
	where we define 
	\begin{equation}\label{G}
		\begin{aligned}
			\nabla G\left( {{{\mathbf{w}}^t}} \right) = \frac{{\sum\nolimits_{n = 1}^N {\sum\nolimits_{i = 1}^{{\beta _n}} {\nabla l\left( {{\mathbf{w}}^t;{\mathbf{x}}_{n,i}^t,y_{n,i}^t} \right)} } }}{{\sum\nolimits_{n = 1}^N {{\beta _n}} }},
		\end{aligned}
	\end{equation}
	and 
	\begin{equation}\label{}
		\begin{aligned}
			\hat F\left( {{{\mathbf{w}}^t}} \right) = \nabla G\left( {{{\mathbf{w}}^t}} \right) - \frac{{\sum\nolimits_{n = 1}^N {a_n^tS_n^t} \sum\nolimits_{i = 1}^{{\beta _n}} {\nabla l\left( {{\mathbf{w}}_n^t;{\mathbf{x}}_{n,i}^t,y_{n,i}^t} \right)} }}{{\sum\nolimits_{n = 1}^N {a_n^tS_n^t{\beta _n}} }}.
		\end{aligned}
	\end{equation}
	
	We can observe from \eqref{G} that $\nabla G\left( {{{\mathbf{w}}^t}} \right) $ denotes the gradient of global FL model with full client participation. 
	In this case, the influence of client selection on the global FL model convergence can be observed, which determines the design of client selection policy.
	In order to make further analysis of FL convergence, we make following assumptions \cite{FL2}:
	\begin{enumerate}[1)]
		\item (Smoothness) The gradient $\nabla G\left( {{{\mathbf{w}}^t}} \right)$ is Lipschitz continuous with respect to ${{\mathbf{w}}^t} $, which can be written as 
		\begin{equation}\label{smooth}
			\begin{aligned}
				\left\| {\nabla G\left( {{{\mathbf{w}}^{t + 1}}} \right) - \nabla G\left( {{{\mathbf{w}}^t}} \right)} \right\| \leqslant L\left\| {{{\mathbf{w}}^{t + 1}} - {{\mathbf{w}}^t}} \right\|,
			\end{aligned}
		\end{equation}
		where $L$ is the Lipschitz constant and $\left\|  \cdot  \right\|$ denotes the norm operator.

		\item (Bound) For any client $n$ and the corresponding sample $i$, the following inequality needs to be satisfied 
		\begin{equation}\label{bound}
			\begin{aligned}
				{\left\| {\nabla l\left( {{\mathbf{w}}_n^t;{\mathbf{x}}_{n,i}^t,y_{n,i}^t} \right)} \right\|^2} \leqslant \rho {\left\| {\nabla G\left( {{{\mathbf{w}}^t}} \right)} \right\|^2},
			\end{aligned}
		\end{equation}
		where $\rho$ is a non-negative constant.
	\end{enumerate}
	Note that, in the FL related literature, the above assumptions are commonly accepted, and widely adopted loss functions are capable of satisfying them.
	Based on this, we can derive Proposition 1.
	
	\noindent {\textbf{Proposition 1}} Given the optimal global FL model $G\left( {{{\mathbf{w}}^ * }} \right)$ and the learning rate $\lambda  = {1 \mathord{\left/{\vphantom {1 L}} \right. \kern-\nulldelimiterspace} L}$, the upper bound of ${{\mathbb{E}}}\left[ {G\left( {{{\mathbf{w}}^{t + 1}}} \right) - G\left( {{{\mathbf{w}}^ * }} \right)} \right]$ at round $t$ can be represented as \cite{FLage}:
	\begin{equation}\label{expect}
		\begin{aligned}
			&	{{\mathbb{E}}}\left[ {G\left( {{{\mathbf{w}}^{t + 1}}} \right) - G\left( {{{\mathbf{w}}^ * }} \right)} \right] \leqslant {{\mathbb{E}}}\left[ {G\left( {{{\mathbf{w}}^t}} \right) - G\left( {{{\mathbf{w}}^ * }} \right)} \right] \\
			&	- \frac{1}{{2L}}{\left\| {\nabla G\left( {{{\mathbf{w}}^t}} \right)} \right\|^2} 
			+ \frac{{2\rho {{\left\| {\nabla G\left( {{{\mathbf{w}}^t}} \right)} \right\|}^2}}}{{L\sum\nolimits_{n = 1}^N {{\beta _n}} }}\sum\limits_{n = 1}^N {{\beta _n}\left( {1 - a_n^tS_n^t} \right)}, \\ 
		\end{aligned}
	\end{equation}
	where ${{\mathbb{E}}}\left[  \cdot  \right]$ is the expectation operator.

	According to Proposition 1, there are three terms that determine the expected gap between the optimal FL model and the global FL model at round $t$.
	In the first term, there is the expected convergence of the global FL model at last round.
	In terms of the second term, it represents the norm of global FL model's gradient at the previous round.
	Note that the third term is bound by the AoU based client selection, where ${a_n^tS_n^t}$ is included.
	In this case, to achieve the minimization of global FL loss, we can minimize the expected gap instead.
	Note that the first and second terms in \eqref{expect} can be regarded as constants.
	Thus, we only need to analyze the third term related to the AoU based client selection.
	By eliminating the postive constant part $\frac{{2\rho {{\left\| {\nabla G\left( {{{\mathbf{w}}^t}} \right)} \right\|}^2}}}{{L\sum\nolimits_{n = 1}^N {{\beta _n}} }}$ and considering the negative sign in the third term, to minimize the expected gap, the optimization problem can be formulated as
	\begin{subequations}\label{max_pro}
		\begin{align}
			\mathop {\max }\limits_{\mathbf{S}} \quad & \sum\limits_{n = 1}^N {a_n^t} {\beta _n}S_n^t \\
			\st\ \quad  & S_n^t \in \left\{ {0,1} \right\},\forall n \in {\cal{N}}. \label{}
		\end{align}
	\end{subequations}
	Problem \eqref{max_pro} is a weighted client selection problem in which both AoU and number of samples play a role in determining the weight.
	Thus, we can select the clients at any round $t$ based on their respective weight ${a_n^t} {\beta _n}$ while ensure constraint \eqref{QoS} is satisfied.
	For instance, the weight of client $n$ can be denoted as ${a_n^t} {\beta _n}$.
	Consequently, the server is able to prioritize clients and order them accordingly, and a selecting list ${\Theta ^t}$ can be obtained which contains all clients at any round $t$ that meets
	\begin{equation}\label{AoU_selection}
		\begin{aligned}
			a_{\left( 1 \right)}^t{\beta _{\left( 1 \right)}} \geqslant a_{\left( 2 \right)}^t{\beta _{\left( 2 \right)}} \geqslant  \cdots  \geqslant a_{\left( N \right)}^t{\beta _{\left( N \right)}}.
		\end{aligned}
	\end{equation}
	In \eqref{AoU_selection}, the expression of $(1)$ denotes the client with highest priority and $(N)$ denotes the client with lowest priority, respectively.
	Note that, if the constraint \eqref{QoS} cannot be met for the selected client $(n)$, the next client $(n + 1)$ will be selected in sequence by the server.
	
	\subsection{Optimization of Resource Allocation}
	Given the AoU based client selection scheme, the original problem \eqref{t_pro} can be transformed to:
	\begin{subequations}\label{t_1}
		\begin{align}
			\mathop {\min }\limits_{{\mathbf{p}},{\mathbf{\tau }}}  \quad &
			\mathop {\max }\limits_{n \in {\cal{N}}^t} \left\{ { {T_n^{cp} + T_n^{cm}} } \right\} \\
			\st\ \quad  &  {R_n^t - {R_s}}  \geqslant 0,  \label{}\\
			& {\tau _n} \in \left[ {0, 1} \right], \label{} \\
			& {p_n^t} \in \left[ {0, {p_n^\text{max}}} \right]. \label{}
		\end{align}
	\end{subequations}
	The above min-max problem is challenging to be solved directly due to its non-convexity.
	However, it can be transformed to a more tractable problem.
	By introducing an auxiliary variable $T$, for any selected client $n \in {\cal{N}}^t $, it is equivalent to transforming the above problem as follows:
	\begin{subequations}\label{t_tran}
		\begin{align}
			\mathop {\min }\limits_{T,{\mathbf{p}},{\mathbf{\tau }}}   \quad &
			T \\
			\st\ \quad  & \frac{{\mu {\beta _n}}}{{{\tau _n}{f_n}}} + \frac{{{d_n}}}{B{\log _2}\left( {1 + \frac{{p_n^t{{\left| {h_n^t} \right|}^2}}}{{\sum\nolimits_{j = n + 1}^{{N^t}} {p_j^t{{\left| {h_j^t} \right|}^2}}  + {\sigma ^2}}}} \right)} \leqslant T, \label{con_T} \\
			&  {{B{\log _2}\left( {1 + \frac{{p_n^t{{\left| {h_n^t} \right|}^2}}}{{\sum\nolimits_{j = n + 1}^{{N^t}} {p_j^t{{\left| {h_j^t} \right|}^2}}  + {\sigma ^2}}}} \right)} - {R_s}}  \geqslant 0,  \label{con_R}\\
			& {\tau _n} \in \left[ {0, 1} \right], \label{con_tau} \\
			& {p_n^t} \in \left[ {0, {p_n^\text{max}}} \right]. \label{con_p}
		\end{align}
	\end{subequations}	
	
	We can easily prove that problem \eqref{t_tran} is convex, as the objective function $T$ and the constraints \eqref{con_T}-\eqref{con_p} are all convex \cite{convex}.
	Based on \eqref{con_T} and \eqref{con_tau}, it can be found that using the maximum CPU frequency coefficient is always efficient, i.e., $\tau _n^* = 1,\forall n \in {\cal{N}}^t$. 
	Because the local training time $T_n^{cp}$ in \eqref{con_T} montonically decreases with $\tau_n$ for any client $n$.
	Thus, to achieve to minimum local training time consumption, the optimal solution of computational resource allocation for problem \eqref{t_tran} can be obtained by $\tau _n^* = 1$.
	
	However, we cannot adopt the same analysis for the power allocation for any client $n$.
	Because in the uplink NOMA transmission, the clients with larger channel gains need to be allocated with higher power so that the interference from other clients decoded behind them can be guaranteed to be small.
	In this case, the performance of NOMA can be guaranteed.
	To derive the optimal power allocation, we first analyze the problem \eqref{t_tran} and then transform it to a more tractable one.
	Note that, to minimize the total time consumption $T$ given the optimal $\tau _n^*$, we can equally maximize the transmission rate of client $n$ based on \eqref{con_T} due to the monotonicity of $T$ with respect to $p_n^t$.
	Moreover, by integrating the QoS constraint \eqref{con_R} into the objective function, we can equally transform the problem \eqref{t_tran} to \cite{TSN}:
	\begin{subequations}\label{R_pro}
		\begin{align}
			\mathop {\max }\limits_{\mathbf{p}} \quad &  \ln \left( {B{{\log }_2}\left( {1 + \frac{{p_n^t{{\left| {h_n^t} \right|}^2}}}{{\sum\nolimits_{j = n + 1}^{{N^t}} {p_j^t{{\left| {h_j^t} \right|}^2}}  + {\sigma ^2}}}} \right) - {R_s}} \right) \\
			\st\ \quad  & {p_n^t} \in \left[ {0, {p_n^\text{max}}} \right]. \label{}
		\end{align}
	\end{subequations}
	
	It can easily be proved that the second derivative of the objective function in \eqref{R_pro} is less than 0, indicating that \eqref{R_pro} is convex.
	Thus, the optimal solution can be obtained by employing the Lagrange dual decomposition approach.
	We can present the Lagrange function of \eqref{R_pro} as
	\begin{equation}\label{L_pro}
		\begin{aligned}
			&{\cal{L} }\left( {{\mathbf{p}}, {\mathbf{v}}} \right)\\
			& = \ln \left( {B{{\log }_2}\left( {1 + \frac{{p_n^t{{\left| {h_n^t} \right|}^2}}}{{\sum\nolimits_{j = n + 1}^{{N^t}} {p_j^t{{\left| {h_j^t} \right|}^2}}  + {\sigma ^2}}}} \right) - {R_s}} \right) \\
			& + \sum\limits_{n \in {{\cal{N}}^t}} {{v_n}\left( {p_n^{\max } - p_n^t} \right)}  \\ 
		\end{aligned},
	\end{equation}
	where ${v_n}$ is the Lagrangian multiplier related to the maximum power constraint.
	Then we have the dual function of problem \eqref{t_tran} as
	\begin{equation}\label{dual_pro}
		\begin{aligned}
			g\left( {{\mathbf{v}}} \right) = \mathop {\max }\limits_{{\mathbf{p}}} {\cal{L} }\left( {{\mathbf{p}},{\mathbf{v}}} \right).
		\end{aligned}
	\end{equation}
	According to the Karush-Kuhn-Tucker (KKT) conditions, we can derive the optimal solution of power allocation in \eqref{t_tran} by taking the partial derivative of ${\cal{L} }\left( {{\mathbf{p}},{\mathbf{v}}} \right)$ with ${\mathbf{p}}$.
	For simplicity of expression, we define ${\phi _n} = 1 + \frac{{p_n^t{{\left| {h_n^t} \right|}^2}}}{{\sum\nolimits_{j = n + 1}^{{N^t}} {p_j^t{{\left| {h_j^t} \right|}^2}}  + {\sigma ^2}}}$, and by the first order derivative of ${\cal{L} }$ with respect to $p_n^t $, we have
	\begin{equation}\label{}
		\begin{aligned}
			\frac{{\partial {\cal{L} }}}{{\partial p_n^t}} = \frac{{B{{\left| {h_n^t} \right|}^2}}}{{\ln 2\left( {\sum\nolimits_{j = n + 1}^{{N^t}} {p_j^t{{\left| {h_j^t} \right|}^2}}  + {\sigma ^2}} \right)}} \\
			\times \frac{{\frac{1}{{{\phi _n}}}}}{{B{{\log }_2}\left( {{\phi _n}} \right) - {R_s}}} - {v_n} = 0.
		\end{aligned}
	\end{equation}
	Furthermore, we define ${\Gamma _n} = {2^{\frac{{{R_s}}}{B}}}$. Then, based on elaborate mathematical analysis, we can derive that
	\begin{equation}\label{}
		{\left( {\frac{{{\phi _n}}}{{{\Gamma _n}}}} \right)^{\frac{{{\phi _n}}}{{{\Gamma _n}}}}} = {2^{{\chi _n}}},
	\end{equation}
	where 
	\begin{equation}\label{}
		{\chi _n} = \frac{{{{\left| {h_n^t} \right|}^2}}}{{\ln 2\left( {\sum\nolimits_{j = n + 1}^{{N^t}} {p_j^t{{\left| {h_j^t} \right|}^2}}  + {\sigma ^2}} \right){v_n}{\Gamma _n}}}.
	\end{equation}
	By the Lambert-W function, ${\frac{{{\phi _n}}}{{{\Gamma _n}}}}$ can be expressed by 
	\begin{equation}\label{}
		\frac{{{\phi _n}}}{{{\Gamma _n}}} = \exp \left( {W\left( {\ln {2^{{\chi _n}}}} \right)} \right),
	\end{equation}
	where $W\left(  \cdot  \right)$ is the expression for the Lambert-W function.
	Substitute $\phi _n$ by ${\phi _n} = 1 + \frac{{p_n^t{{\left| {h_n^t} \right|}^2}}}{{\sum\nolimits_{j = n + 1}^{{N^t}} {p_j^t{{\left| {h_j^t} \right|}^2}}  + {\sigma ^2}}}$, and we can derive the closed-form solution for power allocation as
	\begin{equation}\label{solu_p}
		\begin{aligned}
			p{_n^t }^* = \frac{{\sum\nolimits_{j = n + 1}^{{N^t}} {p_j^t{{\left| {h_j^t} \right|}^2}}  + {\sigma ^2}}}{{{{\left| {h_n^t} \right|}^2}}}\left( {{\Gamma _n}\exp \left( {W\left( {\ln {2^{{\chi _n}}}} \right)} \right) - 1} \right).
		\end{aligned}
	\end{equation}
	Note that ${p _n^t}^*$ contains the power of the clients who are decoded after client $n$.
	In this case, we can first obtain the optimal power of the last decoded client ${N_t}$  as $p{_{N_t}^t }^* = \frac{{{\sigma ^2}}}{{{{\left| {h_{N_t}^t} \right|}^2}}}\left( {{\Gamma _{N_t}}\exp \left( {W\left( {\ln {2^{{\chi _{N_t}}}}} \right)} \right) - 1} \right)$, due to the interference from other clients does not exist for client $N_t$.
	Then, the optimal power expressions of other clients can be derived in sequence according to \eqref{solu_p}.
	Based on problem \eqref{t_tran}, we can obtain the optimal solution of $T$ as follows
	\begin{equation}\label{}
		\begin{aligned}
			{T^ * } = \mathop {\max }\limits_{n \in {{\cal{N}}^t}} \left\{ {T{{_n^{cp}}^*} + T{{_n^{cm}}^*}} \right\},
		\end{aligned}
	\end{equation}
	where $T{_n^{cp} }^*$ and $T{_n^{cm}}^*$ are the results of substituting $\tau _n^ *$ and $p{_n^t }^*$ into $T{_n^{cp} }$ and $T{_n^{cm}}$, respectively.
	
	In order to obtain the Lagrange dual variable ${\mathbf{v}}$, it is necessary to solve the Lagrange dual problem \eqref{dual_pro} expressed as:
	\begin{subequations}\label{}
		\begin{align}
			\mathop {\min }\limits_{{\mathbf{v}}} \quad & g\left( {{\mathbf{v}}} \right)   \\
			\st\ \quad  & {\mathbf{v}} \succeq 0.  
		\end{align}
	\end{subequations}
	Using the subgradient projection method, the Lagrange dual problem can be solved. 
	Specifically, for any selected client ${n \in {{\cal{N}}^t}}$, the subgradient of $g\left( {{\mathbf{v}}} \right)$ can be given by:
	\begin{equation}\label{}
		\begin{aligned}
			\nabla {v_n} = p{_n^t}^ *  - p_n^{\max }.
		\end{aligned}
	\end{equation}
	After deriving the subgradient of ${v_n}$, the Lagrange multiplier ${v_n}$ can be obtained by subgradient projection method iteratively as
	\begin{equation}\label{subgrad}
		\begin{aligned}
			{v_n}\left( {i + 1} \right) = {v_n}\left( i \right) - \eta \nabla {v_n}\left( i \right),
		\end{aligned}
	\end{equation}
	where $i$ and $\eta$ are the iteration index and the step size, respectively.
	
	The procedure of power allocation can be summarized as follows.
	The equation \eqref{subgrad} updates the Lagrangian dual variable ${v_n}$ using the subgradient projection method iteratively.
	This is followed by substituting ${v_n}$ into \eqref{solu_p} to calculate the optimal power allocation $p{ _n^t}^*$.
	Then, the optimal power $p{ _n^t}^*$ is substituted into \eqref{subgrad} to update the Lagrangian dual variable using a cyclic iterative procedure until the convergence is achieved.
	
	\begin{algorithm}[t]
		\footnotesize
		\caption{The Joint AoU based Client Selection and Optimal Resource Allocation Algorithm.}
		\label{algorithm}
		\begin{algorithmic}[1] 
			\STATE \textbf{Initialization:}
			\STATE Initialize variables $\beta_n$, $d_n$, $R_s $, $ {v_n}$ and $A_n^t = 1$ for each client.
			\STATE \textbf{Main Loop:}
			\STATE \textbf{for $n \in  {\cal{N}}$ do} 
			\STATE \quad Obtain $a_n^t$ based on \eqref{a_n};
			\STATE \quad Implement the AoU based client selection scheme based on \eqref{AoU_selection} to obtain the selected clients set $ {\cal{N}}^t$; 
			\STATE \quad \textbf{for $n \in {\cal{N}}^t$ do}
			\STATE \qquad Set $\tau _n^* = 1$ and calculate $p{_n^t }^*$ based on \eqref{solu_p};
			\STATE \qquad Update ${v_n}$ according to \eqref{subgrad};
			\STATE \qquad \textbf{if $R_n^t < R_s$  then}
			\STATE \qquad \quad Delete $n$ from $ {\cal{N}}^t$;
			\STATE \qquad \quad Update $ {\cal{N}}^t$ based on \eqref{AoU_selection};
			\STATE \qquad \textbf{end if}
			\STATE \quad \textbf{end for}
			\STATE \textbf{end for}
		\end{algorithmic}
	\end{algorithm}

	\subsection{Algorithm Design}
	Algorithm \ref{algorithm} summarizes the joint AoU based client selection and optimal resource allocation algorithm.
	Specifically, we firstly initialize the variables $\beta_n$, $d_n$, $R_s $ and $ {v_n}$, and set $A_n^t = 1$ for each client.
	At any FL round $t$, after deriving $a_n^t$ of each client, we perform the AoU based client selection scheme so that the set of selected clients $ {\cal{N}}^t$ can be obtained.
	Subsequently, for each selected client, the optimal solutions for the allocation of computational resource and transmitting power can be calculated based on the monotonicity analysis and dual decomposition method, respectively.
	Note that the set of selected clients will be updated if the transmission quality cannot be guaranteed for any selected client.
	The described main loop in Algorithm \ref{algorithm} will repeat until the convergence of FL.

	\section{Further improvement of FL performance}
	In last section, the AoU based client selection policy is proposed to achieve the minimization of total time consumption via optimal resource allocation at each FL round.
	However, in practical FL scenarios with numerous clients, the client with poor communication resource will not be selected for the global model aggregation. 
	Among these unselected clients, some may have good local FL models, which can contribute more to the global training at the server. 
	To address this issue, in this section, we propose the utilization of the server-side ANN to predict the unselected clients' local FL models at each FL round. 
	Note that ANN outperforms other machine learning techniques when dealing with cluttered and unstructured data, such as images, audio and text.
	It is worthwhile to point out that finding a correlation between diverse clients' local FL models belongs to a regression task, while the ANN has also demonstrated strong capabilities for function fitting tasks \cite{FLT1}.
	In this paper, using ANN, the server can aggregate more clients' local FL models for the  global FL model update, thereby reducing training loss and convergence time.
	
	\subsection{Local FL Models Prediction}
	To obtain the predicting results of the unselected clients' local FL models at any round, the server must utilize a specific client's local model to be the ANN input.
	Hence, one client must be selected to connect with the server and the FL training performance of this client needs to be superior.
	Different from \cite{FLT1} which selects such a client during the entire FL training process, in our scheme, we propose to select this specific client adaptively.
	In particular, we can determine such a client $n^*$ at any round $t$ as follows:
	\begin{equation}\label{}
		{n^ * } = \arg \mathop {\max }\limits_{n \in \cal{N}}{a_n^t} \sum\limits_{i = 1}^{{\beta _n}} {\nabla l\left( {{{\mathbf{w}}^t};{\mathbf{x}}_{n,i}^t,y_{n,i}^t} \right)},
	\end{equation}
	which indicates that client ${n^ * }$ has the highest weight for the FL model aggregation at any round $t$ .
	In the following, we will illusrate the steps of implementing the ANN based method for predicting the local models of clients who are not selected at every FL round.
	
	There are three elements in the ANN-based prediction scheme: a) the input, b) the hidden layer, and c) the output, as described below \cite{ANN}:
	
	{\textbf{The input:}} The input of ANN is the client $n^*$'s local FL model denoted as ${F_{n^*}}\left( {{\mathbf{w}}^t} \right)$, which is used for prediction. As mentioned previously, client $n^*$ is selected at each round adaptively and connects with the server during the whole round. 
	In this case, the input information for the ANN can be provided so that the server can predict the unselected local models.
	
	{\textbf{The hidden layer:}} The hidden layer of ANN can learn the nonlinear relationships between the input ${F_{n^*}}\left( {{\mathbf{w}}^t} \right)$ and the output ${\mathbf{o}}$.
	For simplicity, we only consider a single hidden layer which contains $M$ neurons. 
	We denote the input weight matrix and output weight matrix as ${{\mathbf{v}}^{in}} \in {\mathbb{R }^{M \times K}}$ and ${{\mathbf{v}}^{out}}\in {\mathbb{R }^{K \times M}}$, respectively, where $K$ denotes the number of elements in local FL models.
	
	{\textbf{The output:}} The output of the ANN can be denoted as a vector ${\mathbf{o}} = {F_{{n^*}}}\left( {{{\mathbf{w}}^t}} \right) - {F_n}\left( {{{\mathbf{w}}^t}} \right)$, which indicates the difference of local FL model between client $n^*$ and client $n$.
	In this case, the predicted local FL model of client $n$ can be derived  as ${\hat F_n}\left( {{{\mathbf{w}}^t}} \right) = {F_{{n^*}}}\left( {{{\mathbf{w}}^t}} \right) - {\mathbf{o}}$.
	
	Given the above components of the ANN-based prediction scheme, next we introduce the implementation process to realize the prediction of each unselected client's local FL model.
	In the single hidden layer, the neuronal states can be given by
	\begin{equation}\label{}
		{\mathbf{V}} = \xi \left( {{{\mathbf{v}}^{in}}{F_{{n^*}}}\left( {{{\mathbf{w}}^t}} \right) + {\mathbf{b}}^{in}} \right),
	\end{equation}
	where $\xi \left( x \right) = \frac{2}{{1 + \exp \left( { - 2x} \right)}} - 1$ and ${\mathbf{b}}^{in} \in {\mathbb{R }^{M \times 1}}$ denotes the input bias.
	Based on the above neuronal states, the output of the ANN can be obtained by
	\begin{equation}\label{output}
		{\mathbf{o}} = {{\mathbf{v}}^{out}}{\mathbf{V}} + {\mathbf{b}}^{out},
	\end{equation}
    where ${\mathbf{b}}^{out} \in {\mathbb{R }^{K \times 1}}$ is the output bias.
	Based on \eqref{output}, the prediction result of any unselected client $n$'s local FL model can be calculated as ${\hat F_n}\left( {{{\mathbf{w}}^t}} \right) = {F_{{n^*}}}\left( {{{\mathbf{w}}^t}} \right) - {\mathbf{o}}$.
	
	When the FL local models of unselected clients are predicted via ANN, the global model update at the server now can be rewritten as:
	\begin{equation} \label{global_new}
		F\left( {{{\mathbf{w}}^t}} \right) = \frac{\begin{gathered}
				\sum\nolimits_{n = 1}^N {a_n^{t - 1}S_n^{t - 1}{\beta _n}} {F_n}\left( {{{\mathbf{w}}^{t - 1}}} \right)  \\
				+ \sum\nolimits_{n = 1}^N {a_n^{t - 1}\left( {1 - S_n^{t - 1}} \right){\beta _n}} {\hat F_n}\left( {{{\mathbf{w}}^{t - 1}}} \right){{\mathbbm{1}}_{\left\{ {E_n^t \leqslant \omega } \right\}}}  \\ 
		\end{gathered} }{\begin{gathered}
				\sum\nolimits_{n = 1}^N {a_n^{t - 1}S_n^{t - 1}{\beta _n}}   \\
				+ \sum\nolimits_{n = 1}^N {a_n^{t - 1}\left( {1 - S_n^{t - 1}} \right){\beta _n}} {{\mathbbm{1}}_{\left\{ {E_n^t \leqslant \omega } \right\}}}  \\ 
		\end{gathered} },
	\end{equation}
	where $\sum\nolimits_{n = 1}^N {a_n^{t - 1}S_n^{t - 1}{\beta _n}} {F_n}\left( {{{\mathbf{w}}^{t - 1}}} \right)$ represents the sum of the selected clients' trained local models, and $\sum\nolimits_{n = 1}^N {a_n^{t - 1}\left( {1 - S_n^{t - 1}} \right){\beta _n}} {\hat F_n}\left( {{{\mathbf{w}}^{t - 1}}} \right){{\mathbbm{1}}_{\left\{ {E_n^t \leqslant \omega } \right\}}} $ represents the sum of prediction results of the unselected clients' local FL models at round $t$.
	Denote $E_n^t = \frac{1}{{2K}}{\left\| {{{\hat F}_n}\left( {{{\mathbf{w}}^t}} \right) - {F_n}\left( {{{\mathbf{w}}^t}} \right)} \right\|^2}$ as the prediction error at round $t$ and $\omega$ represents the accuracy requirement for prediction.
	Specifically, in \eqref{global_new}, if it fails to meet the prediction accuracy requirement of the ANN, i.e., ${E_n^t > \omega }$, the prediction result will not be utilized by the server for the global FL model update.
	From \eqref{global_new}, we can observe that, it is possible for the server to obtain additional local FL models by including the prediction results in the model aggregation, hence contributing to a reduction of FL convergence time and training loss.
	Therefore, we can utilize \eqref{global_new} for the model aggregation process to further improve FL performance.

	\subsection{Analysis for ANN based FL}
	\subsubsection{Convergence Analysis} 
	We first assume that ${\left\| {\nabla {F_n}\left( {{{\mathbf{w}}^t}} \right)} \right\|^2} \leqslant \varsigma _1^t + \varsigma _2^t{\left\| {\nabla F\left( {{{\mathbf{w}}^t}} \right)} \right\|^2}$ and $\left\| {\nabla {{\hat F}_n}\left( {{{\mathbf{w}}^t}} \right)} \right\| = \xi _1^t + \xi _2^t\left\| {\nabla F\left( {{{\mathbf{w}}^t}} \right)} \right\|$, where $\left\| {\nabla {{\hat F}_n}\left( {{{\mathbf{w}}^t}} \right)} \right\|$ denotes the gradient bias when predicting the client $n$'s local FL model inaccurately.
	Denote the optimal global FL model as ${{{\mathbf{w}}^ * }}$ which collects the selected clients' local FL models as well as the unselected clients' predicted local FL models.
	In this case, the impact of local model prediction on the global FL model convergence can be observed.
	The following Proposition 2 shows the convergence of the proposed ANN-based FL scheme at any round $t$.
	
	\noindent {\textbf{Proposition 2}} Given the optimal global FL model ${{{\mathbf{w}}^ * }}$, the learning rate $\lambda  = {1 \mathord{\left/{\vphantom {1 L}} \right. \kern-\nulldelimiterspace} L}$, and the gradient bias $\left\| {\nabla {{\hat F}_n}\left( {{{\mathbf{w}}^t}} \right)} \right\|$ caused by predicting inaccuracy, the upper bound of ${{\mathbb{E}}}\left[ {F\left( {{{\mathbf{w}}^{t + 1}}} \right) - F\left( {{{\mathbf{w}}^ * }} \right)} \right]$ at round $t $ can be presented as \cite{FLT1}
	\begin{equation}\label{}
		{{\mathbb{E}}}\left[ {F\left( {{{\mathbf{w}}^{t + 1}}} \right) - F\left( {{{\mathbf{w}}^ * }} \right)} \right] \leqslant \varpi _1^t + \varpi _2^t{{\mathbb{E}}}\left[ {F\left( {{{\mathbf{w}}^{t }}} \right) - F\left( {{{\mathbf{w}}^ * }} \right)} \right],
	\end{equation}
	where $\beta  = \sum\nolimits_{n = 1}^{{{N}}} {{\beta _n}} $, $\varpi _1^t = \frac{{\varsigma _1^t\left( {9\beta  - 8{\mathbb{E}}\left( A \right)} \right)}}{{2L\beta }}$ which determines the convergence of global FL model and ${\mathbb{E}}\left( A \right) = \sum\nolimits_{n = 1}^N {{\beta _n}\left( {1 - S_n^t} \right){{\mathbbm{1}}_{\left\{ {E_n^t \leqslant \omega } \right\}}} \hfill}  + \sum\nolimits_{n = 1}^N {{\beta _n}S_n^t} $, and $\varpi _2^t = \left( {1 - \frac{V}{L} + \frac{{V\varsigma _2^t\left( {9\beta  - 8{\mathbb{E}}\left( A \right)} \right)}}{{L\beta }}} \right)$ also has an impact on the FL convergence time.

	From Proposition 2, we can see that when $\varpi _1^t$ decreases, the gap between ${{{\mathbf{w}}^ t }}$ and ${{{\mathbf{w}}^ * }}$, which can be indicated as the convergence accuracy, also decreases.
	Particularly, when $\varpi _1^t = 0$, with the proposed FL scheme based on ANN, the optimal FL model ${{{\mathbf{w}}^ * }}$ will be obtained, as well as the optimal training loss.
	Besides, from Proposition 2, we can also observe that, when $\varpi _1^t$ and $\varpi _2^t$ reduce, ${{\mathbb{E}}}\left[ {F\left( {{{\mathbf{w}}^{t + 1}}} \right) - F\left( {{{\mathbf{w}}^ * }} \right)} \right] $ will have a lower value than ${{\mathbb{E}}}\left[ {F\left( {{{\mathbf{w}}^{t}}} \right) - F\left( {{{\mathbf{w}}^ * }} \right)} \right] $, indicating that the convergence rate of ${{\mathbf{w}}^ {t+1} }$ to ${{\mathbf{w}}^ * }$ increases.
	Thus, $\varpi _1^t$ and $\varpi _2^t$ jointly determine the FL convergence time.
	Furthermore, Proposition 2 is able to demonstrate that both the learning accuracy and convergence time are affected by the predicting results of local FL models.
	Particularly, when the prediction accuracy requirement is satisfied,
	the predicting local models of unselected clients can be utilized for model aggregation, thereby enhancing the performance of the proposed ANN-based FL scheme.

	\begin{figure}[t]
		\includegraphics[width=0.5\textwidth]{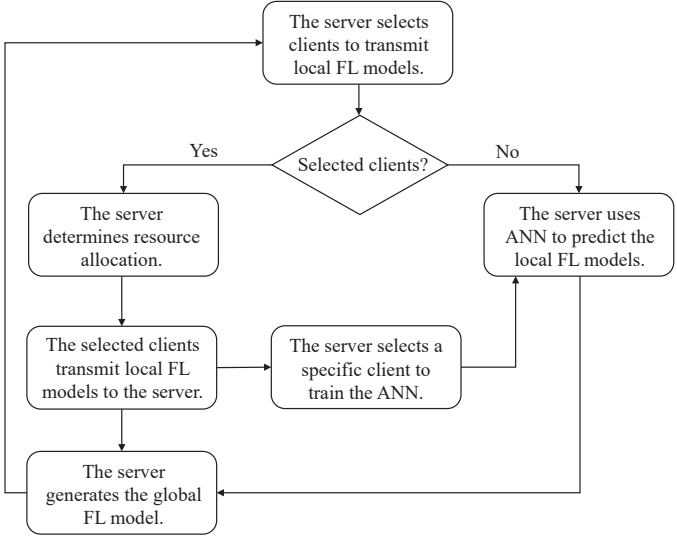}
		\centering
		\caption{The implementation of ANN based FL scheme.}
		\label{Model_ANN_FL}
	\end{figure}

	\subsubsection{Implementation Analysis} There are three steps for the server to make implementation of the proposed ANN-based FL scheme.
	Firstly, the server needs to determine the client selection policy based on the AoU value and data size;
	Secondly, the optimization algorithm should be utilized to realize the resource allocation among selected clients; 
	Lastly, the server must exploit ANN to predict the unselected clients' local FL models.
	More specifically, during each FL round, the server determines the selected clients according to the AoU value and the number of data samples, which is referred as the selecting weight.
	Then, those selected clients send their respective local FL models to the server via uplink NOMA transmission.
	The optimal resource allocation algorithm is performed by the server for each selected client.
	To train the ANN which predicts the unselected clients' local models, the server uses a specific client's local FL model as the ANN input, which owns the highest model aggregation weight at each round.
	Note that, due to the fact that this local FL model was originally used for updating the global FL model, no additional information is required for training the ANN at the server.
	The details of implementation of the proposed ANN based FL scheme can be illustrated in Fig. \ref{Model_ANN_FL}.

	\subsubsection{Complexity Analysis} We firstly need to analyze the complexity of the optimization algorithm for resource allocation.
	Note that at each FL round, the server directly determines the client selection policy according to the AoU value and each client's data size, indicating the complexity of client selection can be regarded as ${\rm O}\left( 1 \right)$.
	Besides, we can obtain the closed-form solutions of computational resource allocation and power allocation for each client.
	Thus, the complexity for resource allocation during one FL round can be regarded as ${\rm O}\left( {{N^t}} \right)$.
	Assume the maximum FL convergence round is $T_m$ and, hence, the complexity for optimization algorithm which obtains the optimal resource allocation during the whole FL convergence time is ${\rm O}\left( {{N^t}} T_m\right)$.
	Moreover, data samples and number of clients determine the complexity of training the ANN.
	Nevertheless, it is assumed that the server has enough computational resources to train the ANN, thus it can be ignored to the overhead of training the ANN.
	
	\begin{table}[t]
		\footnotesize
		\begin{center}
			\caption{\protect\\\textsc{Simulation Parameters}}\vspace{+1em}
			\begin{tabular}{c|c}
				\hline
				\hline
				Parameter & Value\\  \hline
				Radius & $500$ m\\
				Carrier frequency & $1$ GHz\\
				Bandwidth & $B = 1$ MHz \\
				AWGN spectral density & $-174$ dBm/Hz\\
				Path loss exponent & $3.76$\\
				Maximum transmit power & $p_n^\text{max} = 10$  dBm\\
				CPU cycles for each sample & $\mu = 10^7$ \\
				Computation frequency & $f_n = 1$ GHz\\
				Size of local FL model & $d_n = 1$ Mbit\\
				Learning rate & $\lambda = 0.01$ \\
				\hline
				\hline
			\end{tabular}
		\end{center}
	\end{table}

	\section{Simulation Results}
	In this part, the superior performance of the proposed AoU based client selection (ACS) scheme in NOMA network is demonstrated via extensive simulation results.
	A circular network area of radius 500 meters is considered with a server at its center serving 64 uniformly distributed clients.
	We assume a resource constrained scenario, where at most 8 clients can be selected by the server at one FL round, which is far fewer than the total clients.
	The random client selection (RCS) scheme in NOMA and the ACS scheme in OMA are regarded as the benchmarks for performance comparison.
	More specifically, in the RCS scheme, the clients are selected by the server at a random manner at each round, which does not consider the AoU of the clients.
	For the ACS scheme in OMA, the selected clients transmit their trained local FL models via OMA transmission, and the AoU of the clients are considered in this scheme.
	Moreover, as a demonstration of the effectiveness of the proposed ACS with optimal power allocation (ACS-OPA) scheme in NOMA, we compare it with the ACS with random power allocation (ACS-RPA) scheme in NOMA, the RCS with random power (RCS-RPA) scheme in NOMA and the ACS-OPA scheme in OMA.
	Note in these schemes, the same computational resource allocation is adopted and we only consider one channel in the system.
	In our simulation, the standard MNIST data set is used for the FL training.
	Table I presents the simulation parameters.
	
	\subsection{Performance of FL on iid data and non-iid data}
	
	\begin{figure}[t]
		\includegraphics[width=0.5\textwidth]{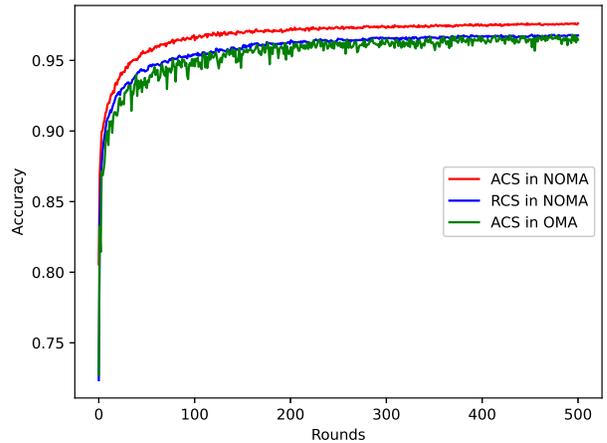}
		\centering
		\caption{FL accuracy on iid data.}
		\label{FL_acc_iid}
	\end{figure}
	
	\begin{figure}[t]
		\includegraphics[width=0.5\textwidth]{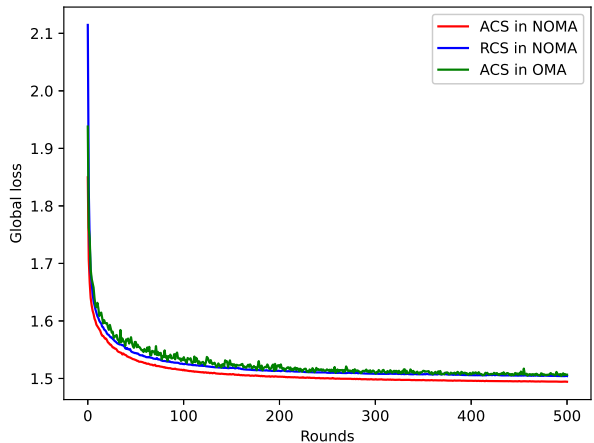}
		\centering
		\caption{FL global loss on iid data.}
		\label{FL_loss_iid}
	\end{figure}
	
	The performance of FL accuracy and global loss on independent and identically distributed (iid) data are shown in Fig. \ref{FL_acc_iid} and Fig. \ref{FL_loss_iid}, respectively.
	It can be found that, with the iid data, the convergence of FL is robust, as there is an identical distribution of labels of MNIST data among all clients.
	From Fig. \ref{FL_acc_iid}, we find that as FL rounds grows, the learning accuracy increases rapidly in 100 rounds and then come to the convergence gradually.
	It can also be found in Fig. \ref{FL_acc_iid} that, compared with other two schemes, our proposed ASC scheme in NOMA can achieve superior learning accuracy performance and hold at the highest learning accuracy value when it reaches the convergence.
	Besides, we observe that the ACS scheme in OMA has the worst robustness, which is because in the OMA based scheme, there are fewer selected clients at each FL round compared with the NOMA based schemes.
	In this case, the FL requires more training rounds so that enough local models can be utilized to achieve the convergence of FL, which is consistent with Fig. \ref{FL_acc_iid}.
	Thus, the effectiveness of NOMA applied in FL can be demonstrated.
	In Fig. \ref{FL_loss_iid}, as FL rounds increase, the global FL loss decreases quickly until achieving the convergence.
	From Fig. \ref{FL_loss_iid}, we also find that the proposed ASC scheme in NOMA can achieve the lowest global FL loss compared with the other two schemes, which validates that the ASC scheme can bring performance improvement in FL systems.
	
\begin{figure}[t]
		\includegraphics[width=0.5\textwidth]{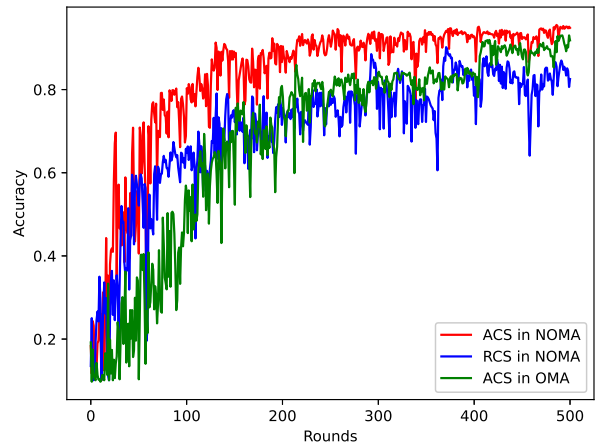}
		\centering
		\caption{FL accuracy on non-iid data.}
		\label{FL_acc_niid}
	\end{figure}

\begin{figure}[t]
		\includegraphics[width=0.5\textwidth]{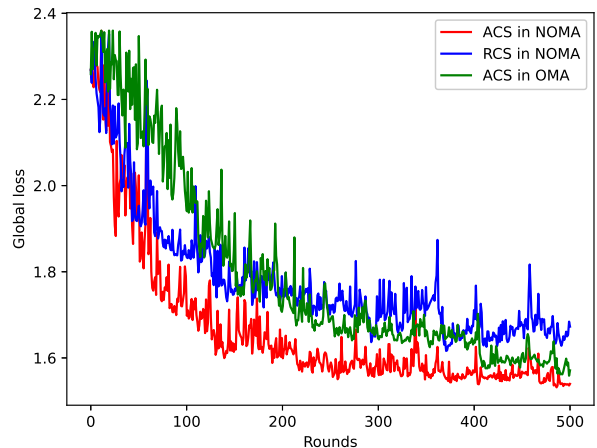}
		\centering
		\caption{FL global loss on non-iid data.}
		\label{FL_loss_niid}
	\end{figure}

	Fig. \ref{FL_acc_niid} and Fig. \ref{FL_loss_niid} present the performance of FL accuracy and global loss on non-iid data, respectively.
	The generation of the non-iid data is based on the label types the clients hold.
	In others words, each client in the system only holds no more than two types of labels, which is a common way to generate the non-iid data \cite{niid}.
	Based on the observation of Fig. \ref{FL_acc_niid} and Fig. \ref{FL_loss_niid}, the overall performance of FL and the robustness of FL convergence decrease a lot on non-iid data compared with that on iid data, which is determined by the property of non-iid data.
	Besides, more FL rounds are required to achieve FL convergence when deploying FL on non-iid data compared with that on iid data.
	Fig. \ref{FL_acc_niid} and Fig. \ref{FL_loss_niid} reveal that our proposed ACS scheme in NOMA can still achieve the highest learning accuracy and the lowest global loss in comparison with the ACS scheme in OMA and the RCS scheme in NOMA.
	This is because with our proposed AoU based scheme, the fairness of client selection in FL can be guaranteed, which is beneficial for the improvement of FL accuracy.
	Besides, the uplink NOMA transmission in our proposed scheme allows the server to select more clients at one FL round, and hence the total data rate is increased and the performance of FL is improved.
	It is also obvious to find that the ACS scheme in OMA has the lowest FL convergence rate when compared with NOMA-based schemes, which is due to the insufficient selected clients at one FL round in OMA based scheme.
	Thus, the effectiveness of NOMA technology in FL can be demonstrated.

	\subsection{Performance of Average AoU and Total Latency}
	\begin{figure}[t]
		\includegraphics[width=0.5\textwidth]{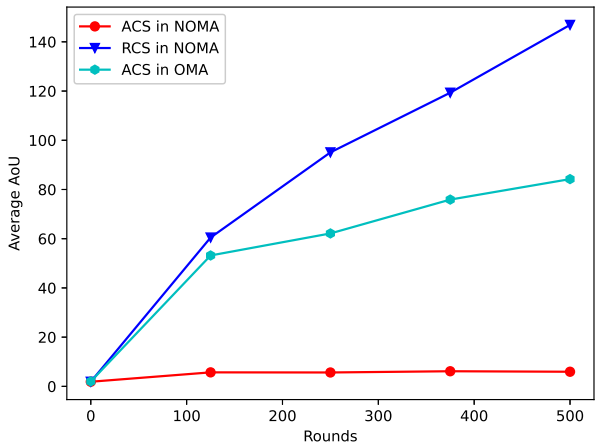}
		\centering
		\caption{Average AoU versus communication rounds.}
		\label{FL_AoU}
	\end{figure}

	Fig. \ref{FL_AoU} shows the performance of the average AoU versus FL communication rounds over the above three schemes.
	Obviously, our proposed ACS scheme in NOMA can achieve the lowest average AoU during the entire process of FL training compared with other two schemes.
	With the growth of FL training rounds, both the average AoU of the RCS scheme in NOMA and the ACS scheme in OMA increase gradually, while the latter has the lower average AoU compared with the former.
	This is because in the ACS scheme in OMA, it also considers the AoU for the client selection, although there is a limited number of clients selected at each FL round.
	Note in our simulation, the selected clients are far fewer than the total clients.
	In this case, for the RCS scheme in NOMA, as the random selecting manner is adopted, the average AoU of the system cannot be guaranteed.
	Nevertheless, the average AoU of our proposed ACS scheme in NOMA can always keep at a relative low value with the increase of communication rounds, i.e., the model update in FL can be guaranteed as fresh as possible, and hence the enhanced FL performance can be realized.
	Besides, the fairness of the client selection in FL can also be guaranteed, which is meaningful in current large scale communication networks consisting of numerous mobile devices.
	
	\begin{figure}[t]
		\includegraphics[width=0.5\textwidth]{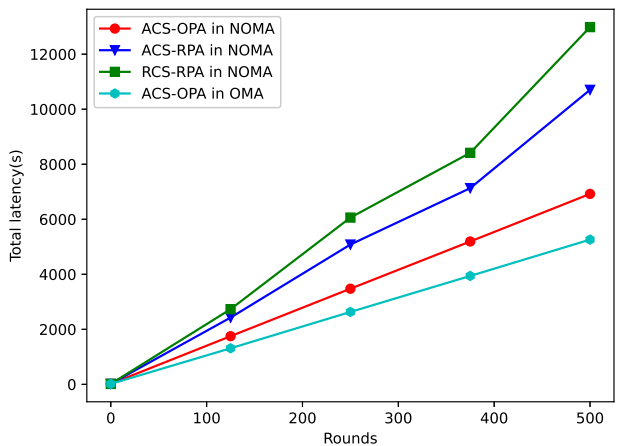}
		\centering
		\caption{Total latency versus communication rounds.}
		\label{Latency_rounds}
	\end{figure}
	
	Fig. \ref{Latency_rounds} shows the total latency performance versus communication rounds. 
	We can observe that in NOMA transmission, compared with the ACS-RPA scheme and the RCS-RPA scheme, our proposed ACS-OPA scheme can achieve the least total latency during the entire FL training process. 
	Thus, our proposed resource allocation scheme can be demonstrated to be effective.
	Besides, we find that the ACS-OPA scheme in OMA can achieve less total latency compared with the NOMA based schemes. 
	This is because the number of selected clients in OMA based scheme is much less than that in NOMA based schemes at each FL round, which leads to less time consumption per round under the synchronous model aggregation criteria. 
	Nevertheless, our proposed NOMA based scheme performs better in terms of FL training accuracy and global loss than the OMA based scheme.

	\subsection{Performance of ANN based FL}
	\begin{figure}[t]
		\includegraphics[width=0.5\textwidth]{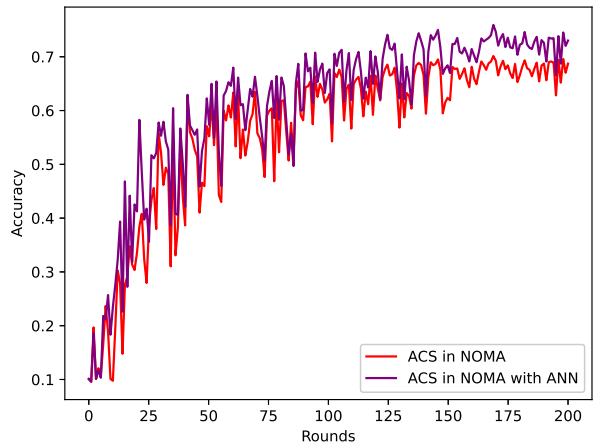}
		\centering
		\caption{ANN based FL accuracy on non-iid data.}
		\label{ANN_acc}
	\end{figure}
    
    \begin{figure}[t]
    	\includegraphics[width=0.5\textwidth]{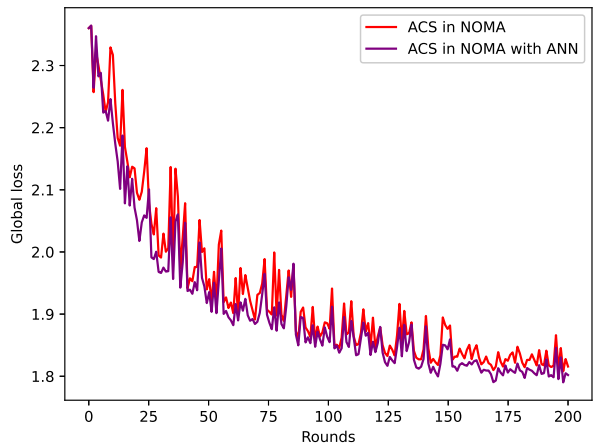}
    	\centering
    	\caption{ANN based FL global loss on non-iid data.}
    	\label{ANN_loss}
    \end{figure}
    
    The learning accuracy and the global loss of ANN based FL on non-iid data are presented in Fig. \ref{ANN_acc} and Fig. \ref{ANN_loss}, respectively, where the ANN is deployed at the server.
    Note that we only consider the scenario of the ANN-based FL on non-iid data here, as it is more meaningful in practical systems, and the similar trend can be found on the iid data.
    It is obvious to find that with the deployment of ANN at the server, both the performance of learning accuracy and global loss of FL can be improved. 
    This is because the deployed ANN allows the server to perform prediction of the local FL models of the  clients who are not selected at each FL round, which equivalently increases the number of local models engaging in the model aggregation without additional wireless communication overhead. 
    Thus, it can be proved that it is effective to deploy ANN at the server side to make further improvement of FL performance.

	\section{Conclusion}
	We proposed an AoU based FL system over NOMA network in this paper, where the effect of AoU on the global FL convergence was analyzed and the communication efficiency was improved via NOMA transmission.
	More specifically, our objective was to achieve the minimization of total time consumption at each FL round, which includes the training time of local FL models and the uplink NOMA transmission time.
	The joint client selection and resource allocation problem was formulated as a non-convex MINLP problem.
	To address it efficiently, we decomposed it into two subproblems, i.e., the client selection subproblem as well as the resource allocation subproblem, and solved them iteratively until FL convergence.
	Particularly, by analyzing the influence of AoU on the global FL loss, we formulated a weighted client selection problem and derived the AoU based client selecting list.
	Given the AoU based client selection scheme, to obtain the optimal resource allocation, we first derived the optimal computational resource allocation based on the monotonic analysis.
	Then, the dual decomposition method was utilized to derive the closed-form solution for power allocation.
	To make further improvement of FL performance, we proposed the use of server-side ANN to achieve the prediction of the unselected clients' local FL models at any round.
	In this case, more local FL models can be utilized for model aggregation at the server, and hence the performance of FL can be improved without extra communication overhead.
	We demonstrated the superior performance of our proposed schemes via extensive simulation results.

	\bibliographystyle{IEEEtran}
	\bibliography{EEref}
	\vspace{0.5em}
	\end{document}